\documentclass[journal]{IEEEtran}
\usepackage[compatibility=false]{caption}
\usepackage{amsmath,amsfonts}
\usepackage{algorithmic}
\usepackage{algorithm}
\usepackage{array}
\usepackage[caption=false,font=normalsize,labelfont=sf,textfont=sf]{subfig}
\usepackage{textcomp}
\usepackage{stfloats}
\usepackage{url}
\usepackage{verbatim}
\usepackage{graphicx}
\usepackage{subcaption}
\usepackage{soul}
\usepackage{booktabs}
\usepackage[table]{xcolor}
\usepackage{cite}
\usepackage{hyperref}
\usepackage{xcolor}
\usepackage[dvipsnames]{xcolor}
\usepackage[inline]{enumitem}
\usepackage{multirow}
\usepackage{capt-of}
\usepackage{blindtext}
\usepackage{tcolorbox}
\usepackage{colortbl}
\usepackage{float}
\begin{document}

\title{ChangeQuery: Advancing Remote Sensing Change Analysis for Natural and Human-Induced Disasters from Visual Detection to Semantic Understanding}
% ChangeQuery: Advancing Remote Sensing Natural and Human-Induced Disaster Change Analysis from Visual Detection to Semantic Understanding
%ChangeQuery: Advancing Change Analysis from Detection to Semantic Understanding
% ChangeQuery: Advancing Change Analysis from Detection to Semantic Understanding for Natural and Human-Induced Disasters

\author{Dongwei Sun,~Jing Yao,~\IEEEmembership{Senior Member,~IEEE},~Kan Wei,~Xiangyong Cao,~Chen Wu,~\IEEEmembership{Member,~IEEE},~Zhenghui Zhao,~Pedram Ghamisi,~\IEEEmembership{Senior Member,~IEEE},~Jun Zhou,~\IEEEmembership{Fellow,~IEEE},~J\'{o}n Atli Benediktsson,~\IEEEmembership{Life Fellow,~IEEE}

% ,~Naoto Yokoya,~\IEEEmembership{Member,~IEEE}
        % <-this % stops a space
\thanks{D. Sun and X. Cao are with the School of Computer Science and Technology and the Ministry of Education Key Lab for Intelligent Networks and Network Security, Xi’an Jiaotong University, Xi’an 710049, China.}
 % (e-mail: sundongwei@outlook.com, caoxiangyong@xjtu.edu.cn)
\thanks{J. Yao and K. Wei are with the State Key Laboratory of Remote Sensing and Digital Earth, Aerospace Information Research Institute, Chinese Academy of Sciences, Beijing, 100094, China.}
 % (e-mail: jasonyao92@gmail.com)
% \thanks{K. Wei is with the Key Laboratory of Target Cognition and Application Technology, Aerospace Information Research Institute, Chinese Academy of Sciences, Beijing 100190, China, and the School of Electronic, Electrical and Communication Engineering, University of Chinese Academy of Sciences, Beijing 100190, China.}
 % (e-mail: weikan24@mails.ucas.ac.cn)
 \thanks{C. Wu and Z. Zhao are with the State Key
Laboratory of Information Engineering in Surveying, Mapping and Remote Sensing, Wuhan University, China.}
%(e-mail:chen.wu@whu.edu.cn, zhaozhenghui@whu.edu.cn)
\thanks{P. Ghamisi is with the Helmholtz-Zentrum Dresden-Rossendorf, 09599 Freiberg, Germany; the Lancaster Environment Centre, Lancaster University, LA1 4YR Lancaster, U.K.; and the Faculty of Electrical and Computer Engineering, University of Iceland, Reykjavík, Iceland.}
 % (e-mail: p.ghamisi@gmail.com)
\thanks{J. Zhou is with the School of Information and Communication Technology, Griffith University, Nathan, QLD 4111, Australia.}
 % (e-mail: jun.zhou@griffith.edu.au)
\thanks{J. A. Benediktsson is with the Faculty of Electrical and Computer Engineering, University of Iceland, 101 Reykjavík, Iceland.}
% (e-mail: benedikt@hi.is).
% \thanks{X. Jia is with the School of Engineering and Technology, University of New South Wales, Canberra, ACT 2612, Australia.}
% (e-mail: x.jia@unsw.edu.au)
}
\markboth{}%
{Shell \MakeLowercase{\textit{et al.}}: A Sample Article Using IEEEtran.cls for IEEE Journals}


\maketitle

\begin{abstract}
Rapid situational awareness is critical in post-disaster response. While remote sensing damage assessment is evolving from pixel-level change detection to high-level semantic analysis, existing vision-language methodologies still struggle to provide actionable intelligence for complex strategic queries. They remain severely constrained by unimodal optical dependence, a prevailing bias towards natural disasters, and a fundamental lack of grounded interactivity. To address these limitations, we present ChangeQuery, a unified multimodal framework designed for comprehensive, all-weather disaster situation awareness. To overcome modality constraints and scenario biases, we construct the Disaster-Induced Change Query (DICQ) dataset, a large-scale benchmark coupling pre-event optical semantics with post-event SAR structural features across a balanced distribution of natural catastrophes and armed conflicts. Furthermore, to provide the high-quality supervision required for interactive reasoning, we propose a novel Automated Semantic Annotation Pipeline. Adhering to a ``statistics-first, generation-later'' paradigm, this engine automatically transforms raw segmentation masks into grounded, hierarchical instruction sets, effectively equipping the model with fine-grained spatial and quantitative awareness. Trained on this structured data, the ChangeQuery architecture operates as an interactive disaster analyst. It supports multi-task reasoning driven by diverse user queries, delivering precise damage quantification, region-specific descriptions, and holistic post-disaster summaries. Extensive experiments demonstrate that ChangeQuery establishes a new state-of-the-art, providing a robust and interpretable solution for complex disaster monitoring. The code is available at \href{https://sundongwei.github.io/changequery/}{https://sundongwei.github.io/changequery/}.
\end{abstract}

\begin{IEEEkeywords}
Remote Sensing Change Detection; Post-Disaster Damage Assessment; Optical–SAR Data Fusion; Vision Language Models
\end{IEEEkeywords}

\section{Introduction}
\IEEEPARstart{T}{he} escalating frequency of catastrophic events, whether triggered by natural forces such as earthquakes and floods or anthropogenic crises like armed conflicts, has underscored the urgency for resilient
monitoring systems\cite{corbane2020big, he2021monitoring, li2024time}. 
\begin{figure}[!t]
    \center
    \includegraphics[width=\linewidth]{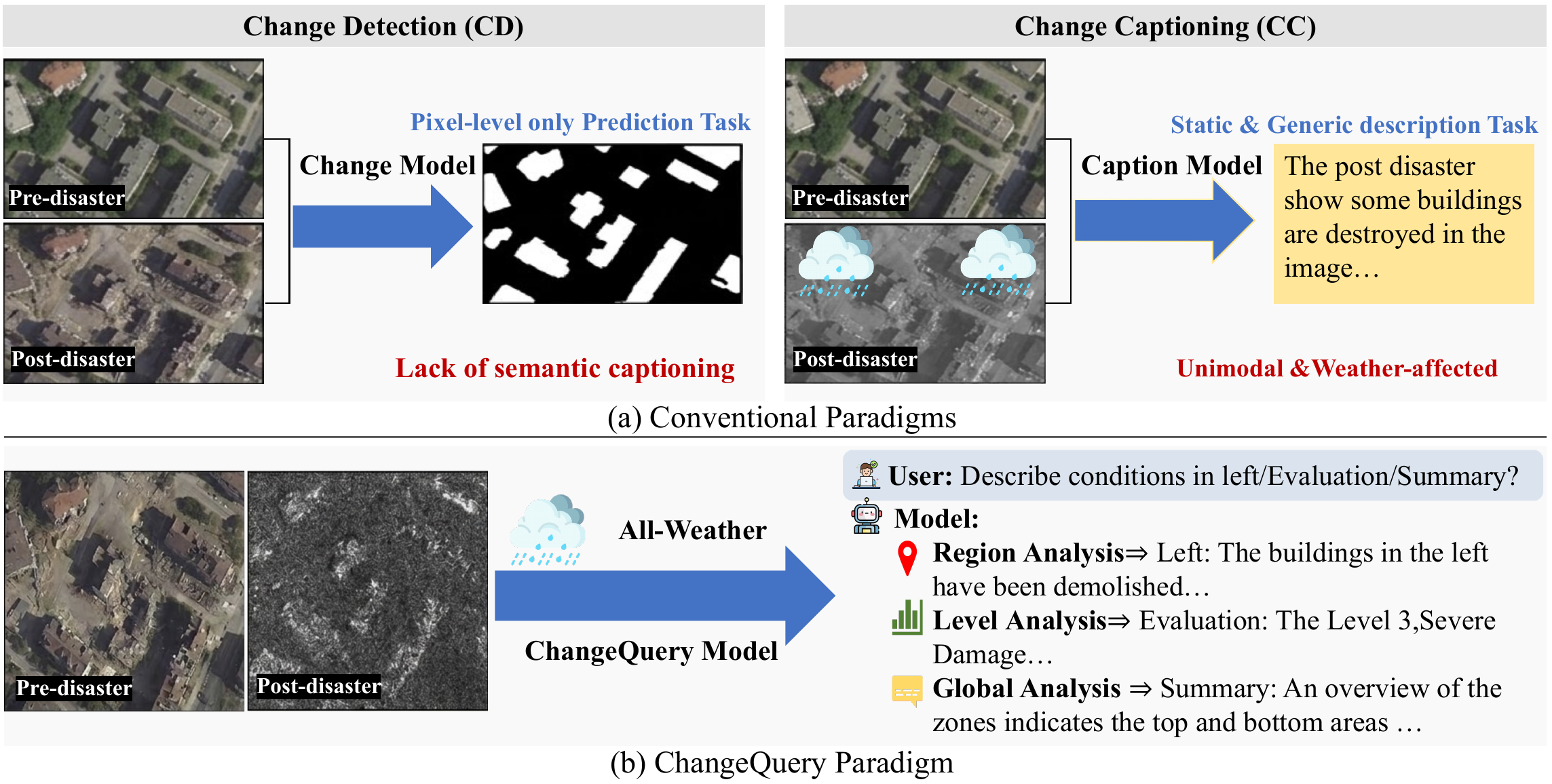}
    \caption{Conceptual comparison of ChangeQuery.
(a) CD yields pixel-level masks without semantics, while RSICC produces generic captions from unimodal (Optical-Optical) inputs.
(b) ChangeQuery adopts a multimodal paradigm (Optical-SAR) for all-weather, multi-granular analysis, including region-level descriptions and hierarchical damage assessment.}
    \label{fig:different_task}
\end{figure}
% For emergency responders, the primary challenge is not merely data acquisition, but rapid situational awareness. Earth Observation (EO) has become the de facto standard for monitoring denied areas \cite{al2024integrating,liuM2CD}. However, the operational pipeline remains heavily fragmented. Traditional CD algorithms typically treat damage assessment as a dense pixel classification problem, outputting binary or categorical masks \cite{Chen2025, Feng2023,10443350}. While visually indicative, these masks represent raw data rather than actionable intelligence. They fail to directly answer high-level strategic questions such as ``\textit{What is the distribution of destroyed structures in the northern sector?}'' or ``\textit{Does the damage pattern suggest localized strikes or widespread inundation?}''
For emergency responders, the primary challenge is not merely data acquisition, but rapid situational awareness. Earth Observation (EO) has become the de facto standard for monitoring denied areas \cite{al2024integrating} \cite{liuM2CD}. However, the operational pipeline remains heavily fragmented. As conceptually illustrated in Fig. \ref{fig:different_task}(a), traditional CD algorithms typically treat damage assessment as a dense pixel classification problem, outputting binary or categorical masks \cite{Chen2025, Feng2023,10443350}. While visually indicative, these masks represent raw data rather than actionable intelligence. They inherently lack the semantic capacity to directly answer high-level questions, such as ``\textit{What is the distribution of destroyed structures in the northern sector?}'' or ``\textit{Does the damage pattern suggest localized strikes or widespread inundation?}''

To inject semantic understanding into this process, early research efforts pivoted towards Remote Sensing Image Change Captioning (RSICC) \cite{hoxha2022change, chang2023changes, liu2024remote}. Yet, as depicted in the conventional captioning paradigm of Fig. \ref{fig:different_task}(a), these models merely generate static, generic sentences. Consequently, they still fail to address the aforementioned complex strategic queries. While modern Visual Language Models (VLMs) \cite{guo2023chat, zhang2024skysense, kuckreja2024geochat} possess the reasoning potential to answer such interactive questions, existing remote sensing VLM approaches exhibit three critical limitations that hinder their deployment in real-world crises:
\begin{enumerate*}[label=(\arabic*)]
    \item \textbf{Modality Constraint:} Most remote sensing VLMs are trained exclusively on optical imagery \cite{zhang2024earthgpt}. This reliance renders them ineffective during the critical 72-hour golden rescue window, which is often plagued by adverse weather, cloud cover, or smoke. SAR, with its active sensing capability, offers a viable solution, yet aligning its non-intuitive backscatter features with semantic language models remains an open challenge \cite{dong2024changeclip}.
    \item \textbf{Scenario Bias:} Current datasets predominantly focus on natural disasters, neglecting the distinct morphological signatures of man-made destruction \cite{weber2020building,wang2025disasterm3}. Unlike the contiguous damage patterns of floods or fires, conflict-induced damage is often discrete, structural, and highly irregular. A robust system must generalize across both natural and anthropogenic domains.
    \item \textbf{Lack of Grounded Interactivity:} Current models lack the granularity to engage in multi-turn reasoning based on physical evidence. They are largely unable to perform precise object counting or oriented localization, which are essential quantitative metrics for logistical planning.
\end{enumerate*}

To address these challenges, we introduce ChangeQuery, a unified framework that redefines disaster analysis as a multimodal, instruction-following task, as illustrated in Fig. \ref{fig:different_task}(b). To support this, we construct the DICQ dataset, a large-scale benchmark that uniquely couples pre-event optical semantics with post-event SAR structural features \cite{stephenson2022deep}. By synthesizing heterogeneous data sources, DICQ covers a spectrum of scenarios ranging from seismic events to urban conflicts, ensuring all-weather robustness. Distinct from previous works that rely on manual captioning or simple template filling \cite{hoxha2022change}, our approach enables the model to generate comprehensive post-disaster change summaries and conduct robust damage evaluation across affected regions, delivering the interactive and granular analysis required for modern disaster response.

To summarize, the contributions of this work are four-fold:
\begin{itemize}
    \item Building upon the BRIGHT benchmark \cite{Chen2025Bright}, we construct the DICQ dataset by integrating independently curated Armed Conflict zones (AC-1 and AC-2) to establish the first heterogeneous Optical-SAR instruction-tuning benchmark. Spanning diverse global locations with varying spatial resolutions (0.3m--1.0m), DICQ ensures a balanced distribution of natural and man-made disasters, fostering the learning of scale-invariant and sensor-agnostic features to effectively mitigate domain shift in all weather analysis.
    % \item We design a novel Automated Semantic Annotation Pipeline that bridges the gap between pixel-level masks and high-level reasoning. By integrating spatially aware zone partitioning, PCA-based oriented localization, and logic-driven damage grading, this framework automatically transforms raw segmentation outputs into grounded, hierarchical instruction sets, enabling precise regional description and quantitative assessment without human intervention.
    \item We propose a novel Automated Semantic Annotation Pipeline adhering to a ``statistics-first, generation-later'' paradigm to provide fine-grained supervision for interactive reasoning. By integrating spatially-aware partitioning, PCA-based localization, and logic-driven grading, it transforms raw masks into grounded instruction sets, effectively endowing the model with precise spatial and quantitative awareness while mitigating hallucinations.
    \item We propose ChangeQuery, a unified multimodal vision language framework that supports multitask reasoning. By integrating a Change-Aware Difference Module with a progressive training strategy, ChangeQuery effectively aligns heterogeneous features with high-level semantic reasoning, establishing a new baseline for interpretable disaster assessment.
    \item Extensive experiments demonstrate that our proposed method achieves State-of-the-Art (SOTA) performance. ChangeQuery consistently surpasses existing generalist VLMs and specialized remote sensing models across multiple metrics, and demonstrates strong cross-domain generalization across heterogeneous disaster types, including both natural and man-made events. This capability establishes a new baseline for interpretable and interactive disaster damage assessment.
\end{itemize}

% The remainder of this paper is organized as follows. Section \ref{sec:related_work} reviews related literature concerning multimodal change detection and remote sensing vision-language models. Section \ref{sec:dataset} details the construction of the DICQ dataset, elaborating on the data composition, the automated semantic annotation pipeline. Section \ref{sec:method} presents the proposed ChangeQuery framework, introducing the overall architecture, the change-aware difference module, and the two-stage training strategy. Section \ref{sec:experiments} reports the experimental results, including quantitative comparisons and qualitative visualizations. Finally, Section \ref{sec:conclusion} summarizes the contributions and conclusion.
The remainder of this paper is organized as follows. Section~\ref{sec:related_work} reviews related work. Section~\ref{sec:dataset} introduces the DICQ dataset and its annotation pipeline. Section~\ref{sec:method} presents the proposed ChangeQuery framework. Section~\ref{sec:experiments} reports experimental results. Finally, Section~\ref{sec:conclusion} concludes the paper.
\section{Related Works}
\label{sec:related_work}
\subsection{Change Detection: From Algebraic to Deep Learning}
Change detection constitutes a fundamental task in remote sensing, aiming to identify distinct differences in the state of a phenomenon by observing it at different times. Historically, this field was dominated by algebraic methods applied to homogeneous optical imagery, such as image differencing, ratioing, and Change Vector Analysis \cite{11184234, bovolo2006theoretical}. With the rapid advancement of deep learning, the paradigm has shifted towards fully convolutional Siamese networks \cite{daudt2018fully, zhang2020deeply}. Recently, Transformer-based architectures have become the standard, utilizing self-attention mechanisms to model long-range spatio-temporal dependencies, thereby improving robustness against seasonal variations and sensor noise \cite{chen2021remote, bandara2022transformer}.

However, in the context of rapid disaster response, the assumption of homogeneous data availability is often violated. Optical sensors\cite{liu2024infrared,zbh,liu2025ctvnet} are frequently rendered ineffective by cloud cover or smoke. Consequently, heterogeneous change detection, particularly the fusion of pre-event optical imagery with post-event SAR data, has emerged as a critical direction. SAR offers all weather imaging capabilities but introduces significant geometric and radiometric distortions compared to optical data \cite{brunner2010earthquake}. To bridge this domain gap, recent methodologies have employed cycle-consistent adversarial networks for image-to-image translation \cite{niu2019conditional, li2021deep} or graph neural networks to align structural features across modalities \cite{sun2022structure, wu2023unsupervised}. Techniques focusing on feature space alignment have also been developed to correlate these distinct modalities for identifying damage patterns in both natural disasters and anthropogenic conflicts \cite{adriano2021learning, wang2022semi}.
Despite these advancements, existing change detection methods predominantly formulate the task as a semantic segmentation problem \cite{zheng2021building, toker2022dynamicearthnet}. The final output is typically a binary or categorical mask. While precise in localization, these masks lack high-level semantic interpretability and cannot explicitly describe the nature of damage or quantify severity in a structured manner useful for strategic decision making.

\subsection{Change Interpretation: From Mask to Language}
To address the semantic limitations of binary masks, the field has gradually expanded towards remote sensing image captioning \cite{lu2017exploring}. Early works utilized encoder-decoder architectures to generate static descriptions of single-temporal scenes. With the advent of bi-temporal analysis, research has shifted towards Change Captioning, which aims to describe differences between two images in natural language. Pioneering works like the RSICCformer \cite{liu2022remote} and attentive fusion networks \cite{chouaf2023captioning,sundw,maskapproxnet,scnet} have demonstrated the feasibility of generating sentences from bi-temporal pairs.

However, these approaches typically function as ``image to text translators" rather than ``intelligent analysts". They are largely limited to homogeneous optical data and produce fixed captions that lack interactivity. Current models struggle to perform grounded tasks such as counting damaged structures, estimating damage ratios, or analyzing specific subregions upon user request. This creates a bottleneck where downstream experts must still manually interpret the generated text to extract operational metrics.

\subsection{Multimodal Learning in Complex Scenarios}
The emergence of General-domain Multimodal Large Language Models (MLLMs), such as LLaVA \cite{liu2023visual} and MiniGPT-4 \cite{zhu2023minigpt}, has revolutionized visual reasoning. By aligning visual encoders with Large Language Models (LLMs), these systems demonstrate impressive zero-shot generalization. However, applying them directly to remote sensing presents unique challenges due to the bird's-eye view perspective, small object scales, and multispectral characteristics \cite{liu2024remoteclip,wang2024earth,hu2023rsgptremotesensingvision}.
\begin{figure*}[!t]
    \center
    \includegraphics[width=\linewidth]{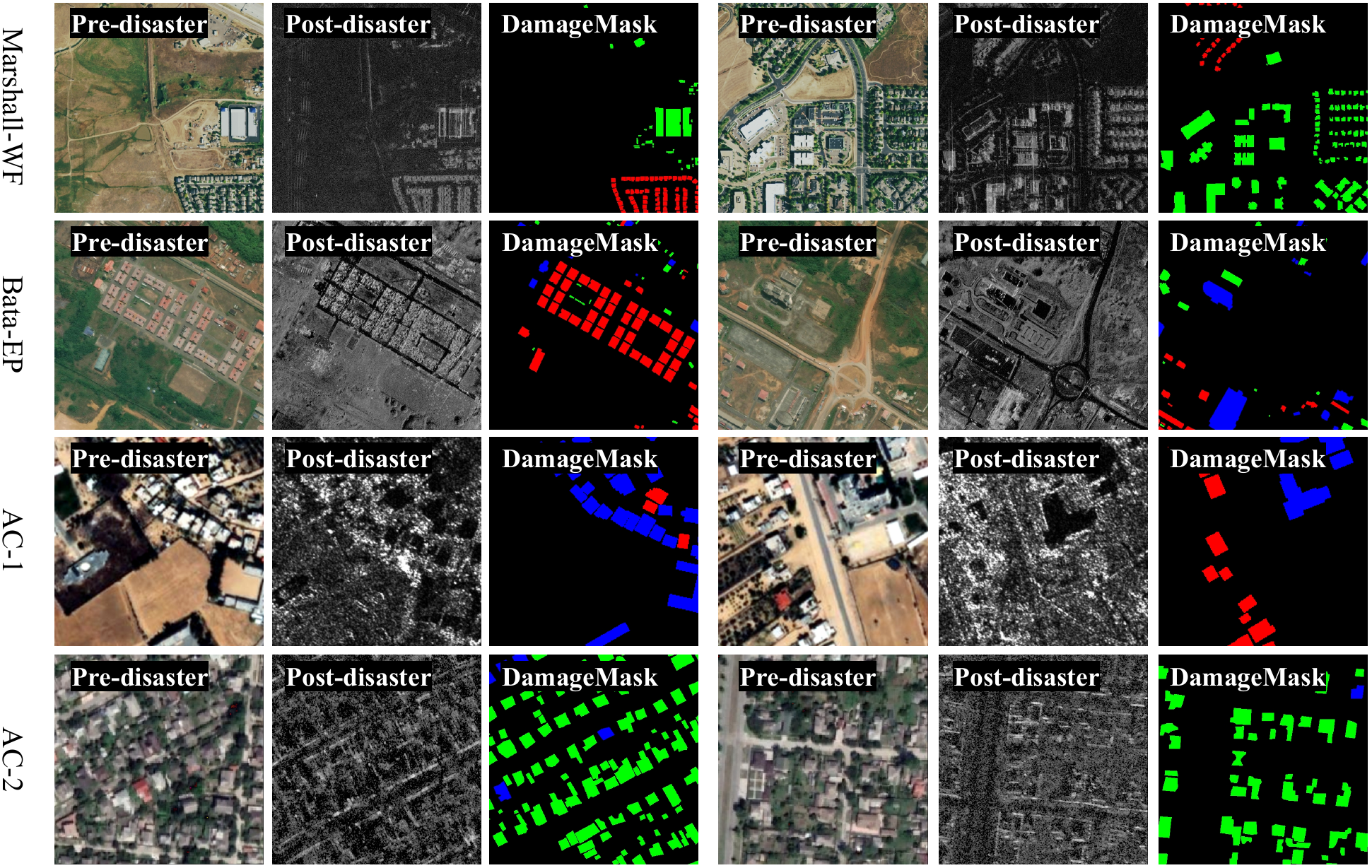}
    \caption{Overview of the DICQ dataset for multi-disaster, multi-source change detection. The dataset covers different geographic locations and disaster types, including wildfire, explosion and armed conflict in Armed Conflict 1(AC-1) and Armed Conflict 2(AC-2) (from top to bottom).
For each disaster scenario, the first column shows the pre-event optical image, the second column shows the post-event SAR image, and the third column displays the corresponding change mask used for supervised learning.
The change mask encodes building-level damage states, where green denotes intact buildings, blue indicates damaged buildings, and red represents destroyed buildings.}
   \label{fig:1}
\end{figure*}
A critical gap in current literature is the handling of complex, non-natural disaster scenarios. While datasets focusing on natural disasters exist, they often neglect the unique signatures of conflict zones, where damage is discrete and structurally complex. Furthermore, existing multimodal fusion techniques in remote sensing often treat SAR as a secondary auxiliary channel rather than a primary source of structural information \cite{10419228}. Our work departs from this by treating optical and SAR as equally critical: optical for pre-event semantic context and SAR for post-event structural verification \cite{mai2023opportunities} \cite{zhan2023rsvg}, thereby enabling a unified framework for all weather, grounded disaster analysis.
\section{Construction of the DICQ Dataset}
\label{sec:dataset}
\subsection{Data Overview and Composition}
The DICQ dataset is meticulously curated to address the limitations of existing benchmarks, which are often restricted to single modalities or specific disaster types. DICQ establishes a large scale, heterogeneous benchmark comprising approximately 70,000 bi-temporal image pairs. A core distinguishing feature of our dataset is its comprehensive typological coverage, as shown in Fig. \ref{fig:1} which categorizes disaster events into three distinct classes: \textit{Natural Disasters(First Row)}, \textit{Man-made Disasters(Second Row)}, and \textit{Conflict Scenarios(Third and Fourth Row)}.
DICQ classifies disaster scenarios into three hierarchical categories to test the model's generalization capabilities in Table
\ref{tab:dataset_overview}:

\textbf{(1) Natural Disasters}: This category represents the foundational component of the dataset (IDs 1-11), encompassing a wide spectrum of meteorological and geological events across diverse terrains. It includes:
\begin{enumerate*}
    \item Seismic Events: Earthquakes in Les Cayes (Haiti), Turkey, Morocco, and Noto (Japan), providing rich samples of structural collapse and debris.
    \item Volcanic Eruptions: Events in Goma (DR Congo) and La Palma (Spain), featuring unique damage patterns caused by lava flow and ash accumulation.
    \item Hydro-meteorological Disasters: Wildfires in Boulder and Maui (USA); Cyclones and Hurricanes in Myanmar (Kyaukpyu) and Mexico (Acapulco); and Floods in Libya (Derna). These scenarios challenge the model to identify inundation and burn scars amidst complex urban backgrounds.
\end{enumerate*}

\begin{table}[!t]
  \centering
  \caption{Overview of disaster events contained in the DICQ dataset. The dataset covers three major categories: Natural Disasters, Man-made Disasters, and Conflicts, with spatial resolutions ranging from 0.3m to 1m.}
    \resizebox{\linewidth}{!}{
    \begin{tabular}{clllc}
    \toprule
    \textbf{ID} & \textbf{Disaster Area} & \textbf{Type} & \textbf{Date} & \textbf{GSD (m)} \\
    \midrule
    \multicolumn{5}{l}{\textbf{Natural Disasters}} \\
    1     & Goma, DR Congo & Volcano Eruption (VE) & May 22, 2021 & 0.33 \\
    2     & Les Cayes, Haiti & Earthquake (EQ) & Aug 14, 2021 & 0.48 \\
    3     & La Palma, Spain & Volcano Eruption (VE) & Sep 19, 2021 & 0.30-0.35 \\
    4     & Boulder, USA & Wildfire (WF) & Dec 30, 2021 & 0.60 \\
    5     & Turkey & Earthquake (EQ) & Feb 06, 2023 & 0.30-0.35 \\
    6     & Kyaukpyu, Myanmar & Cyclone (CC) & May 14, 2023 & 0.60 \\
    7     & Maui, USA & Wildfire (WF) & Aug 08, 2023 & 0.60 \\
    8     & Morocco & Earthquake (EQ) & Sep 08, 2023 & 0.35-0.40 \\
    9     & Derna, Libya & Flood (FL) & Sep 10, 2023 & 0.35 \\
    10    & Acapulco, Mexico & Hurricane (HC) & Oct 25, 2023 & 0.35-0.80 \\
    11    & Noto, Japan & Earthquake (EQ) & Jan 01, 2024 & 0.50 \\
    \midrule
    \multicolumn{5}{l}{\textbf{Man-made Disasters}} \\
    12    & Beirut, Lebanon & Explosion (EP) & Aug 04, 2020 & 1.00 \\
    13    & Bata, Eq. Guinea & Explosion (EP) & Mar 07, 2021 & 0.50 \\
    \midrule
    \multicolumn{5}{l}{\textbf{Conflict Scenarios}} \\
    14    & Armed Conflict Zone 1 & Armed Conflict (AC) & 2022 & 0.60 \\
    15    & Armed Conflict Zone 2 & Armed Conflict (AC) & 2024 & 1.00\\
    \bottomrule
    \end{tabular}%
    }
  \label{tab:dataset_overview}%
\end{table}%
\textbf{(2) Man-made Disasters}: To test the model's ability to detect sudden, high-intensity anthropogenic destruction, we incorporate industrial accidents (IDs 12-13). Specifically, the dataset includes the port explosion in Beirut, Lebanon and the barracks explosion in Bata, Equatorial Guinea. Unlike widespread natural disasters, these events are characterized by concentrated, radial blast damage, requiring precise localization capabilities.

\textbf{(3) Conflict Scenarios}: Uniquely, DICQ extends the scope to armed conflict zones (IDs 14-15). Damage in this category differs fundamentally from natural calamities, exhibiting discrete, irregular, and targeted structural failures rather than contiguous destruction. The inclusion of conflict data fills a critical gap in the remote sensing literature, allowing the model to learn the complex signatures of war-torn infrastructure.

To quantitatively assess the contribution of DICQ, we compare it with mainstream remote sensing image-text datasets in Table \ref{tab:dataset_comparison}. As illustrated, early datasets such as UCM-Captions and RSICD (IDs 1-2) primarily focus on single-image classification with brief descriptions (avg. 12 words). While recent change captioning benchmarks like LEVIR-CC (ID 8) and RSCC (ID 14) have introduced bi-temporal contexts, their textual annotations remain relatively concise (averaging 40 and 72 words, respectively), limiting their utility for complex reasoning tasks.
\begin{table}[!t]
  \centering
  \caption{Comparison of DICQ with existing remote sensing image-text datasets. DICQ significantly outperforms prior benchmarks in terms of semantic density (Average Length), providing rich instruction following data for complex disaster analysis.}
    \resizebox{\linewidth}{!}{
    \begin{tabular}{c|l|c|c|c|c}
    \toprule
    \textbf{ID} & \textbf{Dataset} & \textbf{Year} & \textbf{Image (Pixels)} & \textbf{Captions} & \textbf{Avg. Length} \\
    \midrule
    1     & UCM-Captions & 2016  & 2,100 (1.0B) & 10,500 & 12 \\
    2     & RSICD & 2018  & 10,921 (0.5B) & 54,605 & 12 \\
    3     & fMoW  & 2018  & 1M (437.0B) & N/A   & N/A \\
    4     & SpaceNet 7 & 2021  & 2,389 (2.6B) & N/A   & N/A \\
    5     & S2Looking & 2021  & 5,000 (5.0B) & N/A   & N/A \\
    6     & Qfabric & 2021  & 2,520 (245.1B) & N/A   & N/A \\
    7     & SpaceNet 8 & 2022  & 2,576 (3.0B) & N/A   & N/A \\
    8     & LEVIR-CC & 2022  & 20,154 (1.2B) & 50,385 & 40 \\
    9     & Dubai-CCD & 2022  & 1,000 ($\leq 0.1B$) & 2,500 & 35 \\
    10    & RSICap & 2023  & 2,585 (0.6B) & 2,585 & 60 \\
    11    & VRSBench & 2024  & 29,614 (7.8B) & 29,614 & 52 \\
    12    & WHU-CDC & 2024  & 14,868 (1.9B) & 37,170 & 41 \\
    13    & XLRS-Bench & 2025  & 934 (67.5B) & 934   & 379 \\
    14    & RSCC  & 2025  & 124,702 (32.7B) & 62,351 & 72 \\
    \midrule
    \textbf{15} & \textbf{DICQ (Ours)} & \textbf{2026} & \textbf{136,672} (20B) & \textbf{68,336} & \textbf{571} \\
    \bottomrule
    \end{tabular}%
    }
  \label{tab:dataset_comparison}%
\end{table}%
In stark contrast, DICQ establishes a new standard for semantic density. Although XLRS-Bench (ID 13) offers detailed descriptions, it is constrained by a very small scale (only 934 samples). DICQ strikes a critical balance between scale and depth: it maintains a large volume of 136,672 image pairs while achieving an unprecedented average caption length of 571 words. This figure is nearly 8 times that of RSCC and significantly surpasses all existing benchmarks. This exceptional textual length is driven by our automated annotation engine, which generates not just a single caption, but a structured set of instructions including global summaries, zone-specific descriptions, and quantitative damage assessments. This rich semantic alignment is essential for training MLLMs to perform granular disaster analysis rather than simple image tagging.
\begin{figure*}[htbp]
    \centering
    \includegraphics[width=\linewidth]{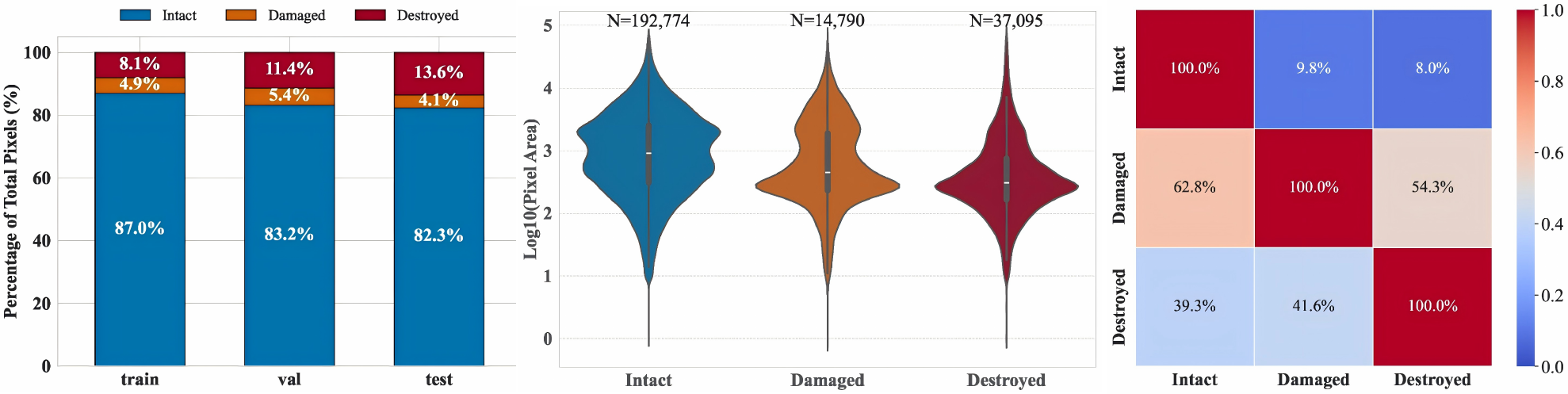}
    \caption{(a) Pixel-wise class balance across the training, validation, and testing splits of the DICQ dataset. The consistent distribution across splits ensures a fair evaluation, despite the natural class imbalance inherent in disaster scenarios. (b) Distribution of building object sizes (in pixel area) across damage categories on a logarithmic scale. The violin plots highlight the multi-scale nature of the DICQ dataset, covering structures from small residential units ($\sim10^2$ pixels) to large industrial complexes ($\sim10^5$ pixels). The instance counts ($N$) further quantify the class imbalance inherent in disaster scenarios. 
    (c) Conditional probability heatmap of class co-occurrence $P(\text{Column} | \text{Row})$. The low probabilities in the first row highlight the spatial isolation of intact zones, while the high values in the second row ($54.3\%$ for Destroyed given Damaged) validate the transitional nature of damage gradients, where partial damage frequently cooccurs with total destruction.}
    \label{fig:all_statistic}
\end{figure*}

We perform a comprehensive statistical analysis to verify the representativeness and physical consistency of the dataset. Regarding the pixel-wise class balance, as illustrated in Fig. \ref{fig:all_statistic}(a), the dataset exhibits a significant class imbalance inherent to real-world scenarios, where the ``Intact'' dominates (82.3\%--87.0\%) over the minority ``Destroyed'' (8.1\%--13.6\%) and ``Damaged'' ($\sim$5\%) categories. Crucially, despite this skew, the distribution pattern remains highly stable across the training, validation, and testing splits. This consistency is vital, ensuring that the test set faithfully reflects the training distribution and provides a rigorous assessment of the model's generalization capability on rare but high-stakes damage classes.

Complementing this pixel-level view, the instance level scale and physics are detailed in Fig. \ref{fig:all_statistic}(b), covering over 240,000 building instances. The counts follow a long tail distribution ($N_{\text{Intact}}=192,774$, $N_{\text{Des}}=37,095$, $N_{\text{Dam}}=14,790$) with vast scale variations ($10^2$--$10^5$ pixels). A key physical insight emerges from the size distribution: ``Damaged'' instances exhibit a larger median pixel area than ``Destroyed'' ones. This aligns with structural physics, where larger, reinforced architectures are more likely to sustain partial structural damage, whereas smaller structures are often completely obliterated in catastrophic events. This multi-scale characteristic requires the model to maintain robust recognition capabilities across varying object resolutions.

% To further quantify spatial context, we analyze the class co-occurrence probability $P(\text{Column} \mid \text{Row})$ in Fig. \ref{fig:all_statistic}(c), revealing two distinct dependency patterns. First, ``Intact'' structures show strong spatial isolation, rarely co-occurring with ``Damaged'' ($9.8\%$) or ``Destroyed'' ($8.0\%$) buildings. This confirms that disaster damage is highly localized, requiring models to strictly suppress false positives in vast safe zones. Second, partial damage exhibits a transitional nature: ``Damaged'' buildings frequently co-occur with ``Destroyed'' ($54.3\%$) and ``Intact'' ($62.8\%$) structures. This reflects real world damage gradients where partially damaged areas serve as complex boundaries between epicenters and safe zones, posing significant challenges for fine-grained semantic disambiguation.
To further quantify spatial context, we analyze the class co-occurrence probability $P(\text{Column} \mid \text{Row})$ in Fig. \ref{fig:all_statistic}(c), revealing two distinct dependency patterns. First, ``Intact'' structures show strong spatial isolation, rarely co-occurring with ``Damaged'' ($9.8\%$) or ``Destroyed'' ($8.0\%$) buildings. This confirms that disaster damage is highly localized. Second, partial damage exhibits a transitional nature: ``Damaged'' buildings frequently co-occur with ``Destroyed'' ($54.3\%$) and ``Intact'' ($62.8\%$) structures. This reflects real world damage gradients where partially damaged areas serve as complex boundaries between epicenters and safe zones, posing significant challenges for fine-grained semantic disambiguation.

\begin{figure}[thbp]
    \centering
    \includegraphics[width=\linewidth]{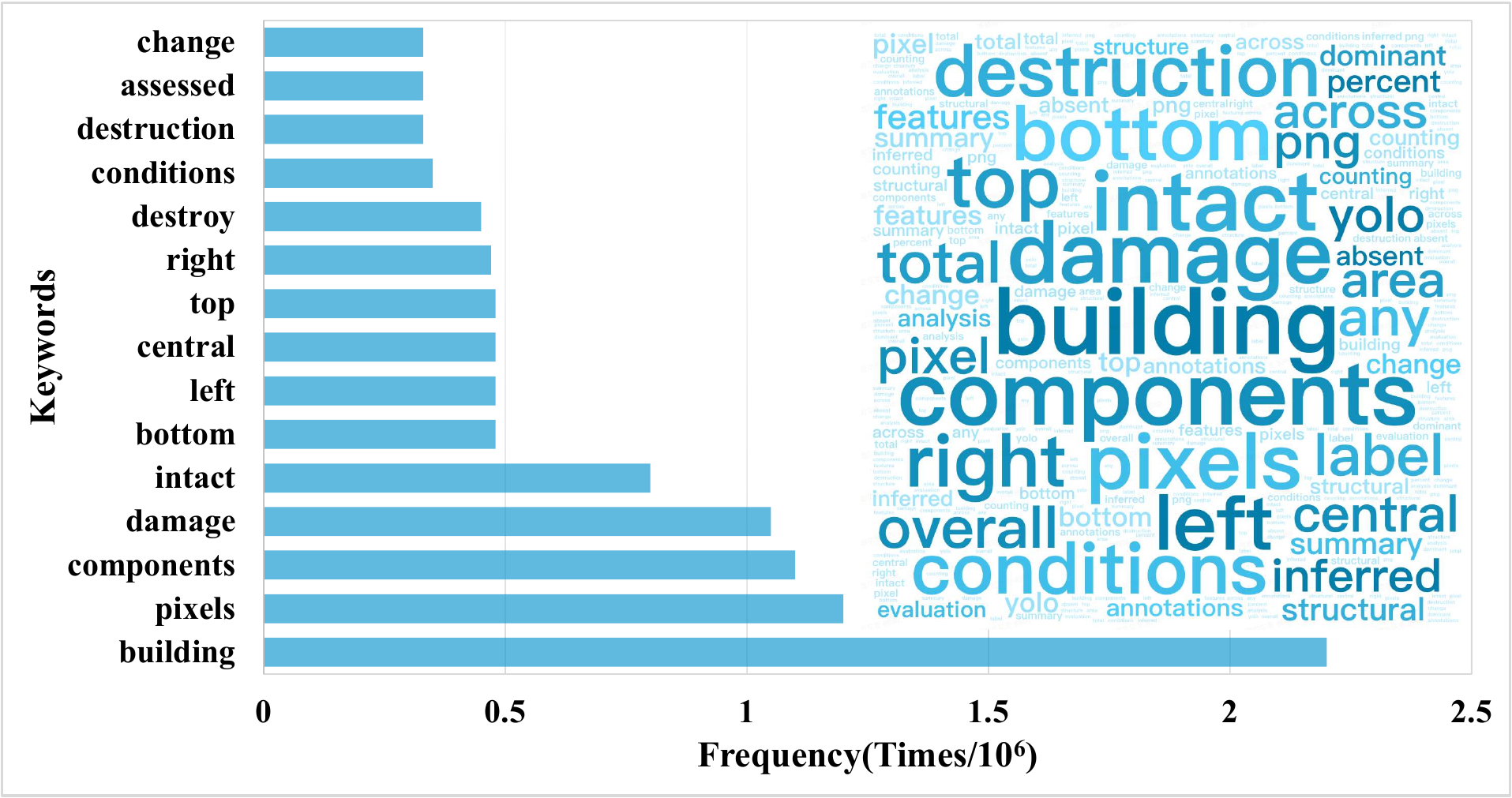}
    \caption{Linguistic analysis of the DICQ dataset. The word cloud highlights the domain-specific focus on building damage assessment. The bar chart reveals a balanced distribution of spatial terms (e.g., top, bottom, central), validating the unbiased nature of our zone based annotation pipeline.}
    \label{fig:word_cloud}
\end{figure}

Finally, we assess the linguistic diversity and spatial balance of the instructions via the word frequency analysis visualized in Fig. \ref{fig:word_cloud}. The vocabulary is highly domain-specific, dominated by terms like ``building'' and ``damage'', ensuring a strict focus on structural integrity. Crucially, the occurrence counts for spatial keywords: ``top'', ``bottom'', ``left'', ``right'', and ``central'' are nearly identical. This uniformity validates our partitioning strategy, ensuring the model receives unbiased supervision across all regions rather than overfitting to the center. Furthermore, the prevalence of fine-grained terms like ``component'' and ``count'' confirms that the instructions contain the precise quantitative details necessary for training rigorous counting and localization tasks.
\subsection{Automated Semantic Annotation Pipeline}
\begin{figure*}
    \centering
    \includegraphics[width=\linewidth]{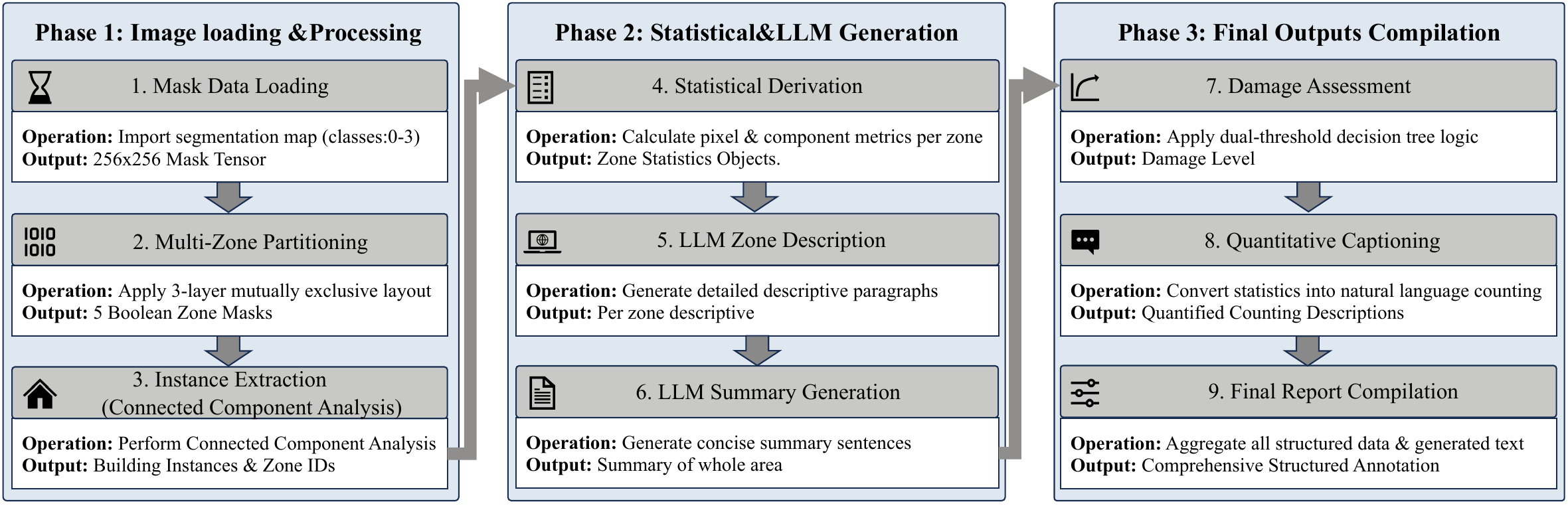}
    \caption{Overview of the proposed Automated Semantic Annotation Pipeline. The workflow transforms raw segmentation masks into comprehensive textual reports through three sequential phases: (1) Image Processing, which performs spatial partitioning and instance extraction via Connected Component Analysis; (2) Statistical Analysis \& LLM Generation, where pixel-level metrics are converted into zone specific descriptions using Large Language Models; and (3) Output Compilation, which aggregates logic driven damage assessments from Level 0 to Level 4 and quantitative summaries into a final structured annotation.}
    \label{fig:pipeline}
\end{figure*}

To systematically bridge the gap between low-level pixel classification and high-level semantic reasoning, we developed an automated processing framework. Let the input segmentation mask be denoted as $\mathbf{M} \in \{0, \dots, C\}^{H \times W}$, where $C=3$ corresponds to the background, intact, damaged, and destroyed classes, indexed from 0 to 3. As illustrated in Fig. \ref{fig:pipeline}, the workflow transforms $\mathbf{M}$ into structured knowledge through a coherent methodology.
\subsubsection{\textbf{Automated Processing Framework}}
\begin{figure*}[t]
    \centering
    \includegraphics[width=\linewidth]{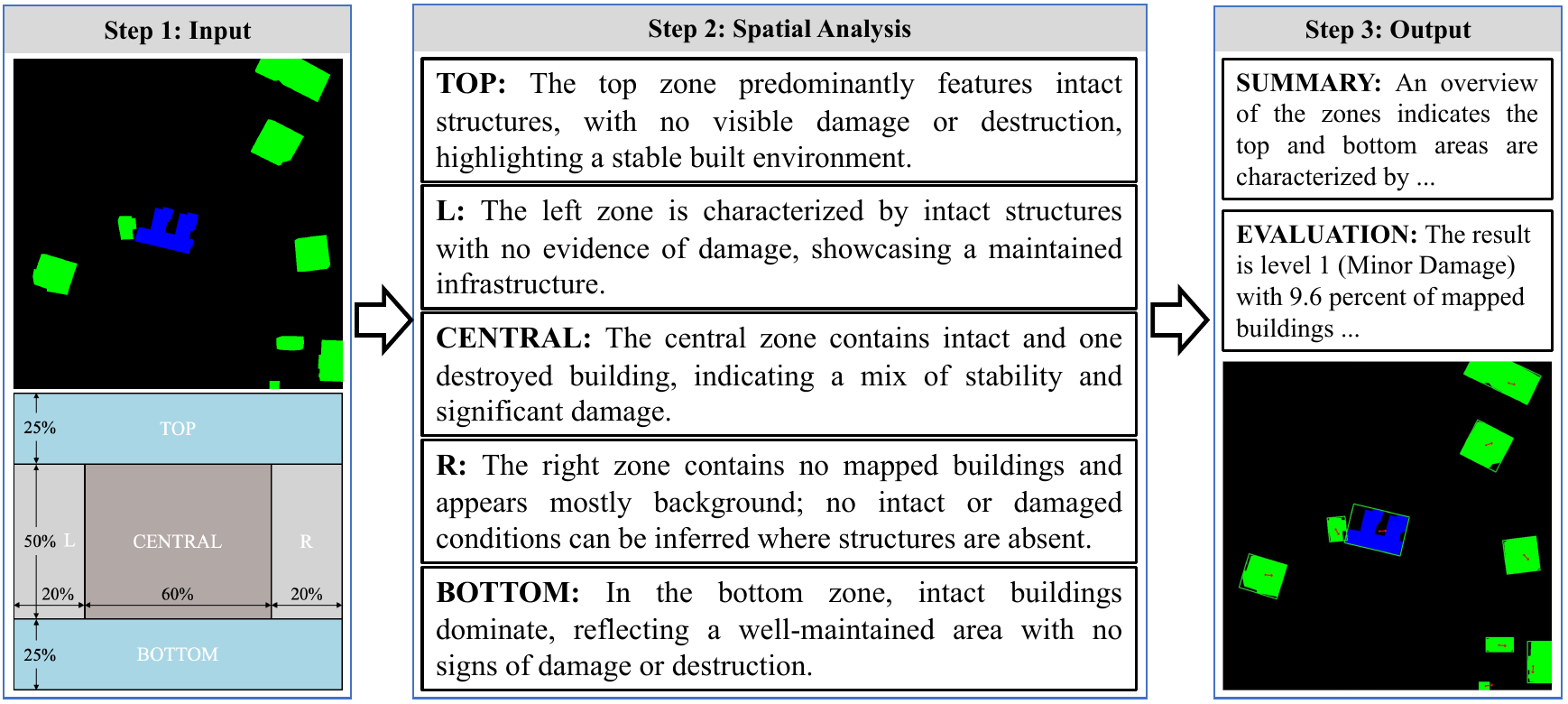}
    \caption{Visualization of the Spatio-Visual Partitioning strategy and sample outputs. To mimic the center-bias tendency of human visual attention, the image domain is strictly partitioned into a dominant Central zone flanked by marginal Top, Bottom, Left, and Right zones. The right panel demonstrates the generated hierarchical annotations, including fine-grained descriptions for each specific zone, a holistic summary, and a quantitative damage evaluation derived from the dual-threshold decision logic.}
    \label{fig:partitioning_framework}
\end{figure*}

The pipeline operates in three sequential phases designed to extract, reason and compile disaster intelligence as in Fig. \ref{fig:partitioning_framework}.
\textit{Phase 1: Spatio-Visual Partitioning and PCA-based Localization.}
The initial phase focuses on geometric parsing. We employ a spatial partitioning function $\mathcal{F}_{zone}$ that maps the image domain $\Omega$ into five mutually exclusive semantic zones $\{Z_{top}, Z_{cen}, Z_{bot}, Z_{left}, Z_{right}\}$. For instance, the dominant $Z_{cen}$ is rigorously defined as:
\begin{equation}
    Z_{cen} = \left\{ (u, v) \mid 0.25H \le v < 0.75H, \ 0.2W \le u < 0.8W \right\}.
\end{equation}
To resolve ambiguities for building instances spanning these boundaries, we adopt a majority-overlap strategy to maintain object integrity. Specifically, an instance $b_i$ is assigned to the zone $z^*$ that contains the maximum number of its constituent pixels $N(b_i, z)$:
\begin{equation}
    z^* = \operatorname*{arg\,max}_{z \in \{Z_{top}, \dots, Z_{right}\}} N(b_i, z).
\end{equation}
Simultaneously, we perform Connected Component Analysis (CCA) to extract the set of building instances. For each instance, we apply Principal Component Analysis (PCA) to derive rotation-aware Oriented Bounding Boxes (OBB). By computing the Singular Value Decomposition (SVD) of the centered coordinates $\bar{\mathbf{X}}_i = \mathbf{U} \mathbf{\Sigma} \mathbf{V}^T$, we obtain the principal orientation $\theta = \arctan(v_{y}/v_{x})$ from the first right singular vector. This formulation provides robust, orientation-invariant localization ($cx, cy, w, h, \theta$), surpassing traditional axis-aligned methods for structures in chaotic debris fields.
\textit{Phase 2: Statistical Derivation and Semantic Reasoning.}
In this phase, we transition from geometric primitives to semantic descriptors. For each zone $z \in \mathcal{Z}$, we compute a dense statistics vector $\mathbf{s}_z \in \mathbb{R}^{C \times 2}$, containing both pixel wise counts and instance wise counts for each damage class. These vectors serve as grounded prompts for the LLM. To mitigate hallucinations, we enforce a ``statistics-first'' constraint: the LLM is tasked to reason explicitly based on $\mathbf{s}_z$ before generating the description. This ensures that the generated text is strictly grounded in physical evidence rather than generative priors. The text generation is modeled as a conditional probability distribution, where the LLM aims to maximize the likelihood of the description $T_z$ given the statistics $\mathbf{s}_z$:
\begin{equation}
    T_z = \arg\max_{T} P_{LLM}(T \mid \text{Prompt}(\mathbf{s}_z, \text{Instruction})).
\end{equation}
\textit{Phase 3: Logic-Driven Damage Assessment.}
% The final phase quantifies the disaster severity into a standardized discrete level $L$ ranging from 0 to 4. We define two critical ratios based on the total building pixels $N_{total}$: the destruction ratio $\rho_{dest} = N_{destroyed} / N_{total}$ and the damage ratio $\rho_{dam} = (N_{damaged} + N_{destroyed}) / N_{total}$. The damage level $L$ is determined by a rigorous dual-threshold decision tree as Eq. \ref{eq:4}. This deterministic logic ensures that the evaluation is objective and consistent across the dataset, regardless of visual variations or subjective biases.
The final phase assigns a disaster severity level $L \in [0,4]$. Given total building pixels $N_{total}$, we compute the destruction ratio $\rho_{dest}=N_{destroyed}/N_{total}$ and the damage ratio 
$\rho_{dam}=(N_{damaged}+N_{destroyed})/N_{total}$. 
To ensure objective and consistent evaluation, the level $L$ is determined by a dual-threshold rule:
\begin{equation}
    L = 
    \begin{cases} 
    4 \ (\text{Destroyed}) & \text{if } \rho_{dest} \ge 0.6 \lor \rho_{dam} \ge 0.85, \\
    3 \ (\text{Severe}) & \text{if } \rho_{dest} \ge 0.3 \lor \rho_{dam} \ge 0.6, \\
    2 \ (\text{Moderate}) & \text{if } \rho_{dest} \ge 0.1 \lor \rho_{dam} \ge 0.35, \\
    1 \ (\text{Minor}) & \text{if } N_{damaged} > 0, \\
    0 \ (\text{No Damage}) & \text{otherwise}.
    \end{cases}
    \label{eq:4}
\end{equation}
This rule ensures that the evaluation is objective and consistent across the dataset, regardless of visual variations.
\subsubsection{\textbf{Structured Annotation and Supervision Signals}}
\begin{figure*}[thbp]
    \centering
    \includegraphics[width=\linewidth]{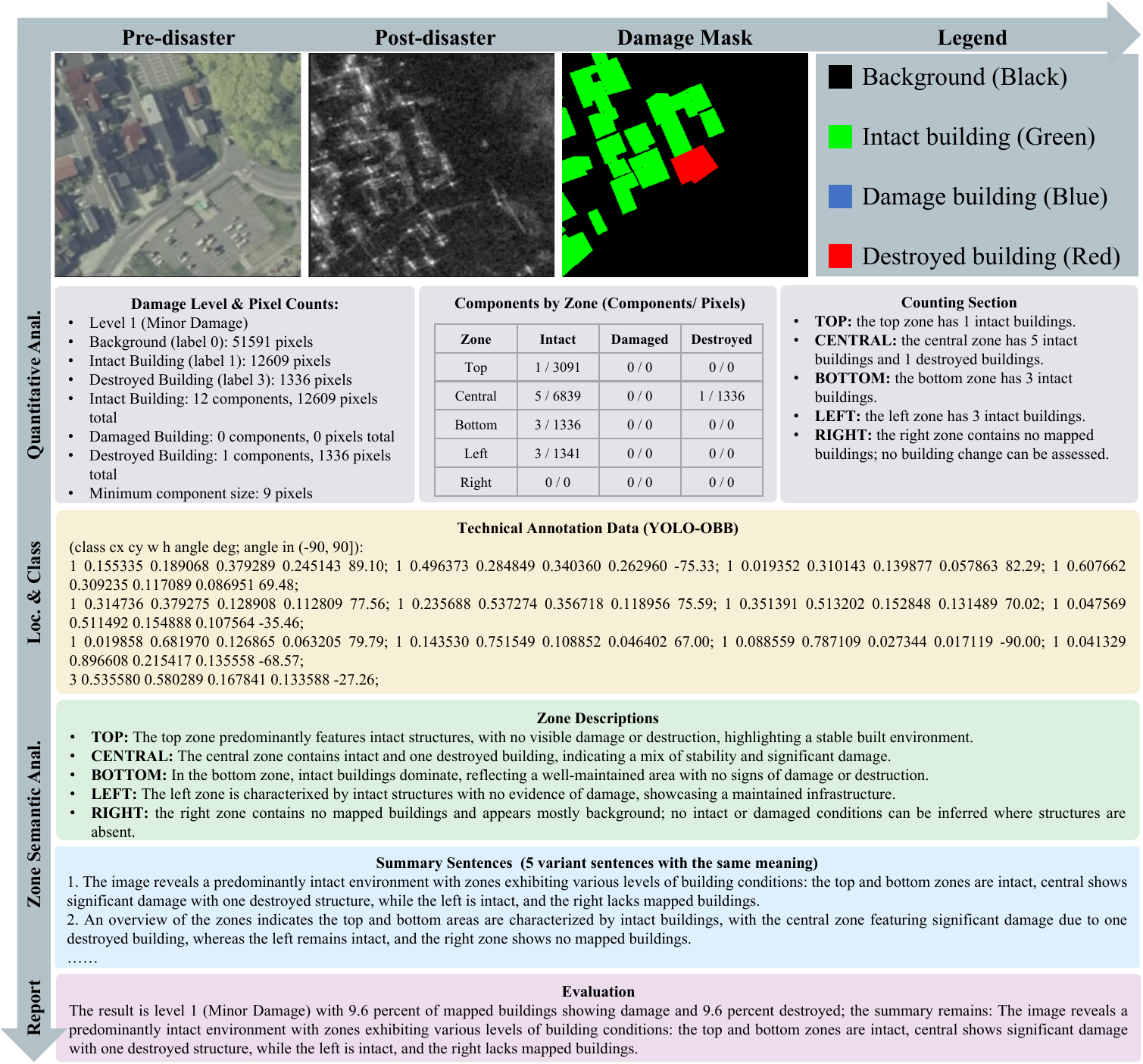}
    \caption{Example of structured annotations from the DICQ dataset generated by our automated pipeline, comprising quantitative statistics including pixel counts and YOLO-OBB coordinates, alongside semantic descriptions with logic-based damage levels, enabling grounded and hallucination-free disaster analysis.}
    \label{fig:annotation_sample}
\end{figure*}

Fig. \ref{fig:annotation_sample} visualizes a representative sample of the final annotation output, which serves as a dense, multimodal knowledge base. As depicted in the quantitative analysis panel, the system provides precise, instance level geometric primitives. The structure components table breaks down building counts by damage status across specific zones, offering fine grained statistical supervision that prevents the model from ignoring small or peripheral objects. Furthermore, the inclusion of YOLO-OBB annotations derived via PCA provides rotation-aware coordinates, including center position, dimensions, and angle. This is particularly critical for disaster scenarios where collapsed structures often exhibit arbitrary orientations that horizontal bounding boxes fail to capture. These mathematical descriptors serve as anchors, ensuring that any subsequent textual generation is strictly grounded in physical reality.

Complementing the numerical data, the semantic analysis panel translates these abstract metrics into coherent natural language. The zone descriptions provide spatially localized narratives, forcing the model to attend to specific image regions such as the top or central zones rather than generating generic captions. Additionally, the counting section explicitly converts the component matrices into natural language quantifiers, directly addressing the numeracy deficit often observed in generalist vision language models. The evaluation module further synthesizes the destruction ratios into a standardized severity rating, simulating the decision making process of a human analyst. By unifying spatial, geometric, and semantic attributes into a structured format, DICQ bridges the semantic gap. This representation enables the model to align visual features with structured logic rather than isolated tokens, providing a foundation for complex reasoning while mitigating hallucinations.
% By encapsulating spatial, geometric, and semantic attributes in this unified structured format, the DICQ dataset effectively bridges the semantic gap. This structured output ensures that the model learns to align visual features not just with isolated words, but with a structured logic system,  providing the foundational bedrock for training models to perform complex reasoning tasks without succumbing to hallucinations.
\section{The ChangeQuery Framework}
\label{sec:method}
\subsection{Overall Architecture}
The ChangeQuery architecture is built upon the LLaVA-1.5 paradigm \cite{liu2023visual}, adapted to accommodate bi-temporal heterogeneous inputs. As illustrated in Fig. \ref{fig:architecture}, the pipeline consists of four primary components: a dual stream visual encoder, a change-aware difference module, a modality aligning projector, and a large language model backbone.
\subsubsection{\textbf{Heterogeneous Visual Encoding}}
The input consists of a bi-temporal image pair: the pre-event optical image $\mathbf{X}_{opt} \in \mathbb{R}^{H \times W \times 3}$ and the post-event SAR image $\mathbf{X}_{sar} \in \mathbb{R}^{H \times W \times C}$. We employ a pre-trained CLIP-ViT as the shared visual encoder $\mathcal{F}_{vis}$. This encoder extracts high-level semantic features from both modalities independently, preserving the distinctive texture of optical data and the backscatter properties of SAR data. The extracted feature are denoted as:
\begin{equation}
    \mathbf{Z}_{opt} = \mathcal{F}_{vis}(\mathbf{X}_{opt}), \quad \mathbf{Z}_{sar} = \mathcal{F}_{vis}(\mathbf{X}_{sar}),
\end{equation}
where $\mathbf{Z} \in \mathbb{R}^{N_v \times D}$ represents the sequence of visual tokens.
\subsubsection{\textbf{Change-Aware Difference Module}}
\begin{figure}[!t]
    \centering
    \includegraphics[width=\linewidth]{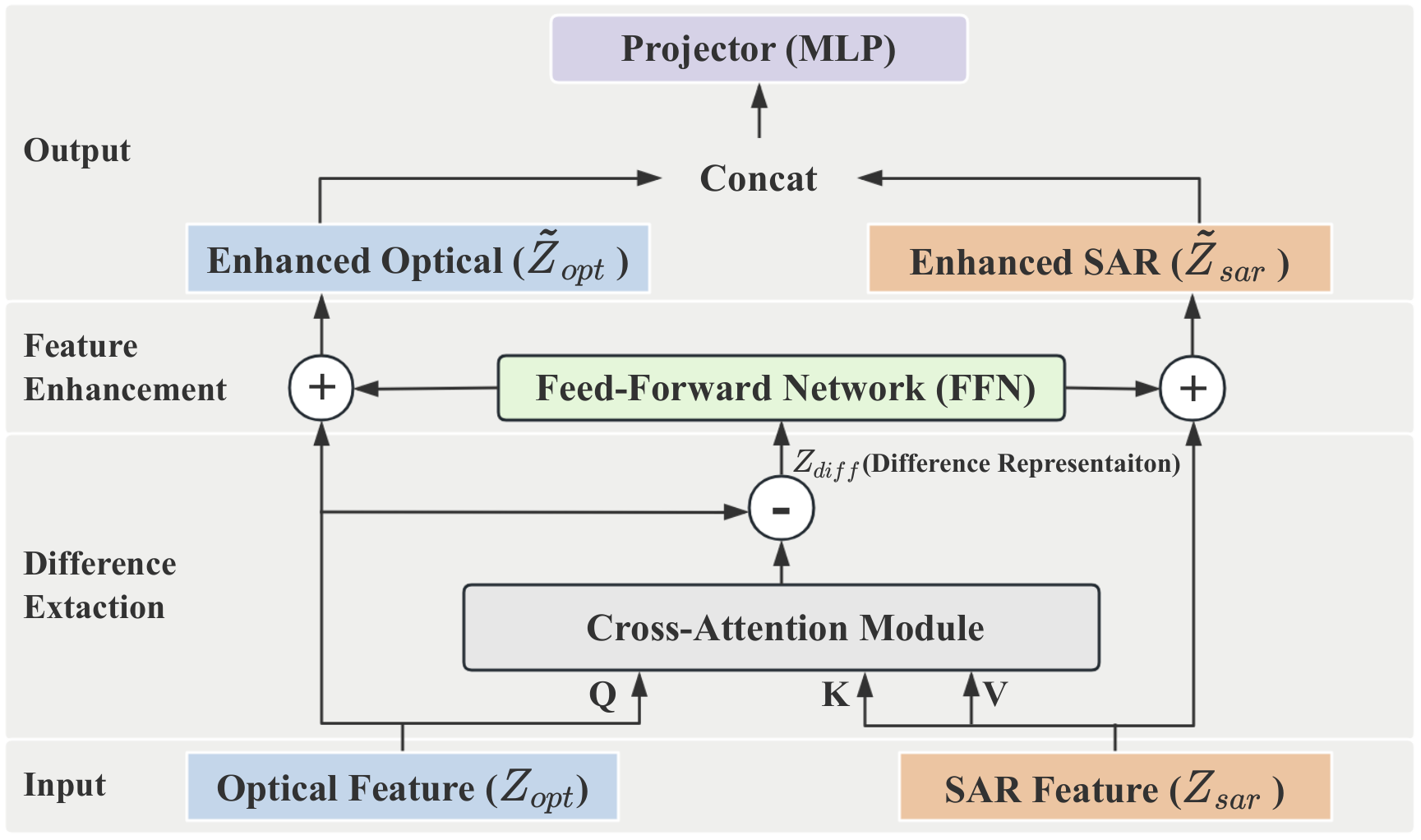}
    \caption{Structure of the Change-Aware Difference Module. It utilizes cross-attention to highlight semantic disparities between pre-event optical and post-event SAR features.}
    \label{fig:diff_module}
\end{figure}

Directly concatenating heterogeneous features often leads to suboptimal performance due to the significant domain gap between optical and SAR modalities. To address this, we introduce a Change-Aware Difference Module inserted between the encoder and the projector. This module operates in two steps: \textit{Difference Extraction} and \textit{Feature Enhancement} in Fig. \ref{fig:diff_module}.

First, to explicitly model the semantic disparity, we employ a Cross-Attention mechanism. We treat the optical features as queries ($Q$) to attend to the SAR features (Keys $K$ and Values $V$). This aligns the post-event structural information with the pre-event semantic context, yielding an explicit difference representation $\mathbf{Z}_{diff}$:
\begin{equation}
    \mathbf{Z}_{diff} = \text{CrossAttn}(Q=\mathbf{Z}_{opt}, K=\mathbf{Z}_{sar}, V=\mathbf{Z}_{sar}) - \mathbf{Z}_{opt}.
    \label{eq:diff_extract}
\end{equation}
Here, the subtraction operation highlights the features in the SAR imagery that deviate from the optical baseline, effectively isolating the change signal.

Second, to ensure the visual tokens passed to the LLM are sensitive to these changes, we perform feature injection. We utilize $\mathbf{Z}_{diff}$ to enhance the original modality-specific features via a residual update mechanism. This generates the final change-aware representations $\tilde{\mathbf{Z}}_{opt}$ and $\tilde{\mathbf{Z}}_{sar}$:
\begin{equation}
    \tilde{\mathbf{Z}}_{opt} = \mathbf{Z}_{opt} + \alpha \cdot \text{FFN}(\mathbf{Z}_{diff}), \quad \tilde{\mathbf{Z}}_{sar} = \mathbf{Z}_{sar} + \beta \cdot \text{FFN}(\mathbf{Z}_{diff}),
    \label{eq:feature_enhance}
\end{equation}
where $\alpha$ and $\beta$ are learnable gating factors and FFN is Feed-Forward Network. This step injects the extracted change information back into the contextual stream, ensuring that both modalities are aware of the damage patterns before projection.
\subsubsection{\textbf{Visual Language Projection}}
To bridge the gap between frozen visual features and the language embedding space, we utilize a Multi Layer Perceptron (MLP) as the trainable connector $\mathcal{P}$. Unlike single-image architectures, our projector handles the concatenated sequence of the bi-temporal features. We fuse the optical and SAR features along the sequence dimension to form a unified visual context.
\begin{equation}
    \mathbf{H}_{vis} = \mathcal{P}([\tilde{\mathbf{Z}}_{opt} ; \tilde{\mathbf{Z}}_{sar}]),
\end{equation}
where $[;]$ denotes concatenation and $\mathbf{H}_{vis} \in \mathbb{R}^{2N_v \times D_{llm}}$ represents the projected visual tokens aligned with the LLM's word embedding space. This early fusion strategy allows the model to implicitly learn the correlation and differences between the two time steps via the self attention of the LLM.
\subsubsection{\textbf{LLM Reasoning Backbone}}
% For the reasoning core, we employ Vicuna-7B v1.5, a decoder-only transformer fine-tuned on instruction-following data. The model takes a sequence of inputs comprising the visual tokens $\mathbf{H}_{vis}$ and the textual instruction tokens $\mathbf{H}_{instr}$ derived from the user query (e.g., ``Assess the damage in the central zone").
% The generation of the response $\mathbf{Y}$ is formulated as an auto-regressive prediction problem. The probability of generating the target response tokens $y_t$ is conditioned on the multimodal context and previously generated tokens:
% \begin{equation}
%     p(\mathbf{Y} \mid \mathbf{X}_{opt}, \mathbf{X}_{sar}, \mathbf{X}_{instr}) = \prod_{t=1}^{L} p(y_t \mid \mathbf{H}_{vis}, \mathbf{H}_{instr}, y_{<t}).
% \end{equation}
% During training, we keep the visual encoder frozen to retain its generalized feature extraction capability, while the MLP projector and the LLM backbone are finetuned using LoRA to adapt to the specific domain of disaster assessment.
For the reasoning core, we adopt Vicuna-7B v1.5, a decoder-only Transformer fine-tuned for instruction-following tasks. The model processes a multimodal sequence composed of visual tokens $\mathbf{H}_{vis}$ and textual instruction tokens $\mathbf{H}_{instr}$ derived from the user query (e.g., ``Assess the damage in the central zone'').
The response generation is formulated as an autoregressive process. Specifically, the probability of generating the target sequence $\mathbf{Y}$ is defined as:
\begin{equation}
    p(\mathbf{Y} \mid \mathbf{X}_{opt}, \mathbf{X}_{sar}, \mathbf{X}_{instr}) = \prod_{t=1}^{L} p(y_t \mid \mathbf{H}_{vis}, \mathbf{H}_{instr}, y_{<t}).
\end{equation}
During training, the visual encoder is kept frozen to preserve its general representation capability. In contrast, the MLP projector and the LLM backbone are fine-tuned using LoRA, enabling efficient adaptation to the disaster domain.
\begin{figure*}[!ht]
    \centering
    \includegraphics[width=\linewidth]{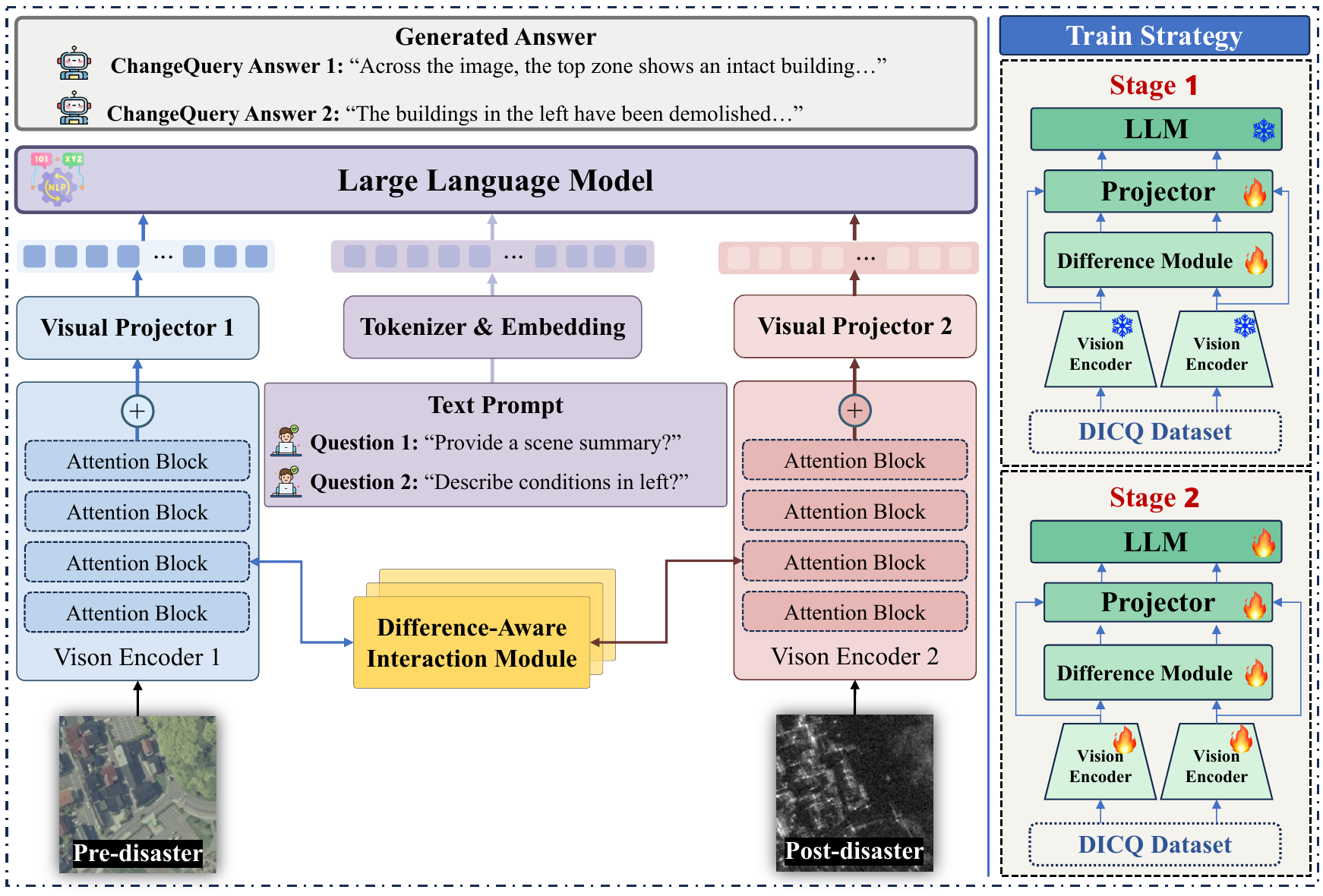}
    \caption{The overall architecture of ChangeQuery. The model takes a pre-event optical image and a post-event SAR image as inputs. Features are extracted via a shared CLIP-ViT encoder, concatenated, and projected into the language space. The Vicuna-7B LLM then processes these visual tokens alongside the text instruction to generate grounded responses. The right part is to show our train strategy of changequery model.}
    \label{fig:architecture}
\end{figure*}
\subsection{Two-Stage Training Strategy}
Training a Multimodal Large Language Model on heterogeneous remote sensing data presents significant optimization challenges, primarily due to the distinct feature distributions of optical and SAR modalities and the risk of catastrophic forgetting in the pre-trained backbone. To ensure stable convergence and robust instruction following, we adopt a progressive training paradigm, as illustrated in Fig. \ref{fig:architecture}, which decouples feature alignment from semantic reasoning.
\subsubsection{Stage 1: Modality Alignment and Feature Initialization}
In the initial phase, the primary objective is to bridge the semantic gap between the visual encoders and the LLM while initializing the change detection capability. During this stage, we keep the massive parameters of both the dual stream Vision Encoders and the LLM Backbone frozen. Optimization is restricted exclusively to the Difference Module and the MLP Projector. This strategy is motivated by the need to construct a stable visual language interface before burdening the model with complex reasoning tasks. Since the Difference Module is initialized from scratch, this focused training allows it to learn the fundamental mapping correlations between optical texture and SAR backscatter without perturbing the generalized knowledge inherent in the pre-trained backbones. Effectively, this stage forces the connector layers to act as a translator, converting raw heterogeneous feature disparities into coherent linguistic embeddings.
\subsubsection{Stage 2: End-to-End Instruction Tuning}
Subsequently, the training transitions to a full scale instruction tuning phase designed to master the specific syntax of disaster reporting. In this stage, we unfreeze the Vision Encoders to allow for domain adaptation, enabling the CLIP model to adjust to the nadir perspective of remote sensing imagery. Simultaneously, the LLM Backbone is finetuned using LoRA.
Crucially, this stage is driven by the structured multi-turn conversations generated by our pipeline. The training targets are not merely static captions but interactive dialogues requiring hierarchical reasoning. The model is explicitly supervised to perform tasks ranging from global damage grading (e.g., ``Level 0'') to fine-grained regional descriptions. This rigorous supervision ensures the model learns to ground its textual outputs in the specific quantitative evidence extracted from the heterogeneous visual inputs, effectively minimizing hallucinations.
\section{Experiments and Analysis}
\label{sec:experiments}
\subsection{Experimental Settings}
\textbf{Baselines.}
To rigorously evaluate the performance of ChangeQuery, we benchmark it against two categories of state-of-the-art vision-language models:
\begin{enumerate*}[label=(\arabic*)]
    \item Generalist MLLMs: We compare with leading open-source multimodal models that support multi-image or interleaved inputs, including LLaVA-NeXT-Interleave (8B) \cite{li2024llava}, LLaVA-OneVision (8B) \cite{libo2024llava}, and InternVL 3 (8B) \cite{zhu2025internvl3}. These models represent the current state-of-the-art in general visual reasoning capabilities.
    \item Specialized Remote Sensing Models: We also include domain-specific change captioning models, specifically CCExpert \cite{wang2024ccexpert} and TEOChat \cite{irvin2024teochat}, to assess the advantages of our proposed heterogeneous fusion strategy against existing remote sensing solutions.
\end{enumerate*}

\textbf{Implementation Details.}
The ChangeQuery model is implemented using the PyTorch framework. For the visual encoder, we utilize the pre-trained CLIP-ViT, keeping its weights frozen to preserve generalized feature extraction. The MLP projector and the Vicuna-7B v1.5 backbone are fine-tuned using LoRA \cite{hu2021lora} to efficiently adapt to the disaster assessment domain. We utilize the AdamW optimizer with a cosine learning rate scheduler. To handle the heterogeneous inputs, optical and SAR images are resized to $252 \times 252$ pixels before encoding.
\subsection{Evaluation Metrics}
Evaluating the quality of generated disaster reports requires assessing both the textual overlap with ground truth and the semantic accuracy of the content. We employ two sets of metrics:

\textbf{(1) N-Gram Overlap Metrics.}
Following standard practices in image captioning, we report ROUGE-L \cite{lin2004rouge} and METEOR \cite{banerjee2005meteor}. These metrics measure the precision and recall of n-grams between the generated response and the reference text. While they provide a basic assessment of fluency and lexical overlap, they are often insufficient for capturing the complex reasoning and structural details required in long form disaster analysis.

\textbf{(2) Semantic Similarity Metric (ST5-SCS).}
To address the limitations of n-gram metrics in capturing semantic alignment for long-form descriptions, we adopt Sharpened Cosine Similarity (ST5-SCS). Utilizing the Sentence T5 encoder \cite{ni2022sentence} for embedding extraction, the score is computed as:
\begin{equation}
    \text{SCS}(u, v) = \text{Cosine}(u, v)^3 \cdot \text{sgn}(\text{Cosine}(u, v)),
\end{equation}
where $u$ and $v$ represent the embeddings. 

Quantitative results in Table \ref{tab:performance_comparison} reveal distinct performance gaps among the evaluated methods. Generalist Models, led by InternVL 3 (8B), achieve a respectable semantic consistency (50.46\% SCS) due to their robust pre-trained backbones. However, they suffer from suboptimal N-gram scores (e.g., ROUGE $\approx$ 10\%), indicating a failure to grasp the specific lexicon of disaster reporting and the complex backscatter characteristics of SAR imagery. While Remote Sensing(RS) specialized models like CCExpert improve upon this by adapting to aerial views, achieving a competitive SCS of 50.68\%, they still lack explicit mechanisms to model the structural disparities between optical and SAR modalities. Consequently, they struggle to generate the precise, fine-grained details required by the DICQ benchmark, limiting their fluency and factual accuracy.

In contrast, ChangeQuery significantly outperforms all baselines, establishing a new SOTA. Most notably, it achieves a METEOR score of 22.00\%, surpassing the second-best method (RSCC) by a substantial margin of over 8 points. Since METEOR correlates strongly with human judgment on harmonic mean of precision and recall, this leap indicates that our model generates reports with far superior coherence and logical structure. Furthermore, the highest SCS of 54.07\% confirms that ChangeQuery effectively bridges the modality gap. We attribute these gains to the Change-Aware Difference Module, which aligns heterogeneous features, and Two-Stage Instruction Tuning, which strictly grounds the textual generation in physical evidence rather than generic hallucinations.
\begin{table*}[t]
\centering
\caption{Quantitative comparison with state-of-the-art methods on the DICQ benchmark. We categorize baselines into \textit{Generalist MLLMs} (trained on natural images) and \textit{Remote Sensing (RS) Specific Models} (fine-tuned on aerial data). Evaluation metrics include N-gram overlap (ROUGE, METEOR) for textual fluency and Sharpened Cosine Similarity (SCS) for semantic consistency. The best results are highlighted in \textbf{bold}, demonstrating that ChangeQuery achieves superior performance across all dimensions.}
\label{tab:performance_comparison}
\setlength{\tabcolsep}{10pt} 
\begin{tabular}{lcccccc}
\toprule
\multirow{2}{*}{\textbf{Method}} & \multirow{2}{*}{\textbf{Params}} & \multirow{2}{*}{\textbf{Base Model}} & \multirow{2}{*}{\textbf{Reference}} & \multicolumn{2}{c}{\textbf{N-Gram}} & \textbf{Contextual Similarity} \\
\cmidrule(lr){5-6} \cmidrule(lr){7-7}
& & & & ROUGE (\%) \ensuremath{\uparrow} & METEOR (\%) \ensuremath{\uparrow} & ST5-SCS (\%) \ensuremath{\uparrow} \\
\midrule
\multicolumn{7}{l}{\textit{\textbf{General Model}}} \\
\midrule
LLaVA-NeXT-Interleave & 8B & Qwen 1.5 & arXiv'24 \cite{li2024llava} & 10.13 & 14.08 & 45.11 \\
LLaVA-OneVision       & 8B & Qwen 3   & arXiv'24 \cite{libo2024llava} & 9.71  & 13.30 & 46.07 \\
InternVL 3            & 8B & Qwen 2.5 & arXiv'25 \cite{zhu2025internvl3} & 10.62 & 12.19 & 50.46 \\
\midrule
\multicolumn{7}{l}{\textit{\textbf{RS Specific Model}}} \\
\midrule
CCExpert              & 7B & Qwen 2   & arXiv'25 \cite{wang2024ccexpert} & 10.55 & 13.00 & 50.68 \\
TEOChat               & 7B & Llama 2  & ICLR'25 \cite{irvin2024teochat}  & 8.16  & 9.21  & 46.22 \\
RSCC                  & 7B & Qwen 2.5 & NeurIPS'25 \cite{chen2025rscc} & 10.67 & 13.71 & 47.21 \\
\midrule
\rowcolor{gray!10}
\textbf{ChangeQuery} & \textbf{7B} & \textbf{Vicuna 1.5} & \textbf{Ours} & \textbf{12.91} & \textbf{22.00} & \textbf{54.07} \\
\bottomrule
\end{tabular}
\end{table*}
\begin{figure*}[!t]
    \centering
    \includegraphics[width=\linewidth]{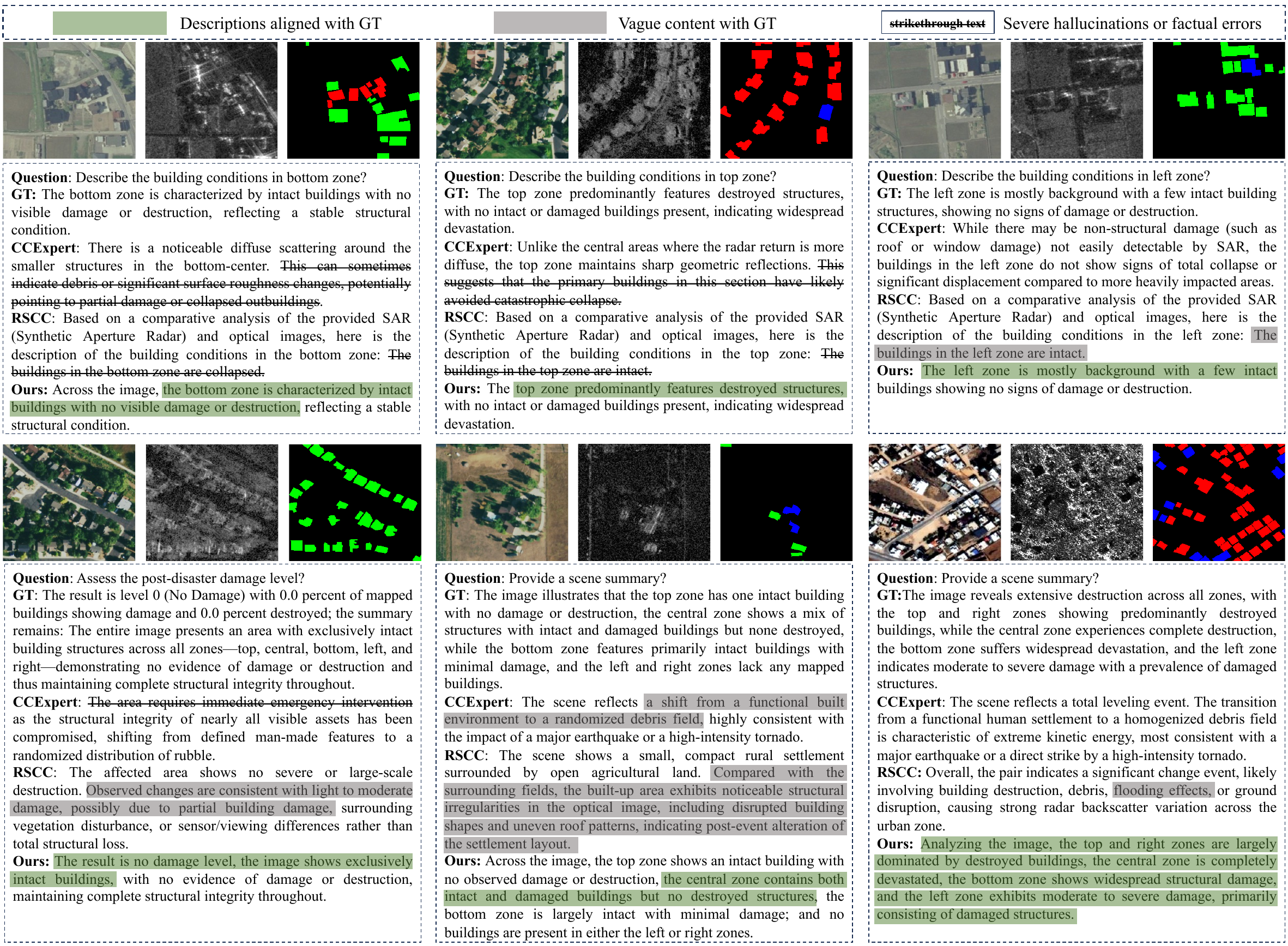}
    \caption{Qualitative comparison of diverse instruction-following tasks on the DICQ benchmark. We compare the responses generated by \textbf{ChangeQuery} against state-of-the-art baselines including \textbf{CCExpert} and \textbf{RSCC}. To facilitate visualization, accurate descriptions aligned with the Ground Truth (GT) are highlighted in \textbf{\textcolor{ForestGreen}{green}}, while severe hallucinations or factual errors are marked with \textbf{\st{strikethrough text}}, and vague content is highlighted in \textbf{\textcolor{gray}{gray}}. The examples illustrate that ChangeQuery generates more grounded and spatially precise analysis, successfully avoiding the common issues of disaster bias and cross-modal misalignment observed in baseline methods.}
    \label{fig:expshow}
\end{figure*}
\subsection{Qualitative Visualization \& Analysis}
Fig. \ref{fig:expshow} presents a visual qualitative comparison between ChangeQuery and state-of-the-art baselines, including CCExpert and RSCC, across diverse disaster scenarios. To facilitate analysis, descriptions aligned with the ground truth are highlighted in \textbf{\textcolor{ForestGreen}{green}}, whereas severe hallucinations or factual errors are marked with \textbf{\st{strikethrough text}}, and vague or inaccurate descriptions are highlighted in \textbf{\textcolor{gray}{gray}}. Three critical observations can be drawn from these results:

\textbf{(1) Mitigation of ``Disaster Bias'' and Hallucinations.}
A prevalent issue in existing remote sensing VLMs is the tendency to hallucinate damage even in intact areas, a phenomenon we term ``disaster bias''. This is clearly observable in the bottom-left case which represents a Level 0 scenario with no damage. Both CCExpert and RSCC misinterpret the scene, fabricating descriptions of ``immediate emergency intervention'' or ``light to moderate damage'' despite the buildings being structurally sound. This error likely stems from their inability to distinguish SAR speckle noise from actual debris. In contrast, ChangeQuery correctly identifies the scene with ``exclusively intact buildings'', demonstrating superior robustness against false positives in complex SAR environments.

\textbf{(2) Accurate Cross Modal Alignment for Destruction.}
The top middle example focusing on the top zone illustrates a scenario of severe destruction. While the ground truth indicates ``widespread devastation'', CCExpert erroneously claims the zone maintains sharp geometric reflections and avoided catastrophic collapse, while RSCC incorrectly labels it as intact. These failures highlight the struggle of baseline models to align pre-event optical semantics with post-event SAR scattering patterns. Empowered by the proposed Difference Module, ChangeQuery accurately captures the loss of structural coherence in the SAR imagery and correctly describes the zone as predominantly featuring destroyed structures.

\textbf{(3) Precise Spatial Reasoning and Granularity.}
The scene summary tasks presented in the bottom-middle and bottom-right panels test the ability of the models to perform zone level spatial reasoning. The baselines often generate generic, holistic descriptions such as CCExpert's generalized claim of a ``randomized debris field'' or hallucinate specific disaster types like tornadoes or flooding without visual evidence. Conversely, ChangeQuery exhibits precise spatial grounding. As demonstrated in the bottom-middle instance, it successfully disentangles complex mixed scenarios by accurately identifying intact and damaged buildings in the central zone while noting minimal damage in the bottom zone. This fine-grained descriptive capability validates the effectiveness of our zone-aware instruction tuning strategy.

In summary, ChangeQuery not only generates more accurate damage assessments but also significantly reduces the hallucination rate compared to specialized baselines, providing reliable and grounded intelligence for disaster response.
\section{Conclusion}
\label{sec:conclusion}
In this paper, we introduced ChangeQuery, a unified multimodal framework that redefines disaster damage assessment by shifting the paradigm from low level pixel classification to high-level semantic reasoning. Addressing the critical limitations of existing methodologies, we constructed the DICQ dataset. This large scale benchmark uniquely couples pre-event optical semantics with post-event SAR structural features, covering a diverse spectrum of scenarios ranging from natural catastrophes to anthropogenic conflicts. By employing a novel automated semantic annotation pipeline, we transformed raw segmentation masks into grounded, hierarchical instruction sets, enabling the model to learn spatial reasoning and precise quantification without human intervention.

Methodologically, the ChangeQuery framework incorporates a Change-Aware Difference Module and utilizes a progressive two-stage training strategy to effectively bridge the significant domain gap between heterogeneous modalities. Extensive experiments demonstrate that our approach achieves SOTA performance across multiple metrics, significantly outperforming both generalist vision language models and specialized remote sensing baselines. The model exhibits remarkable capabilities in generating spatially precise logic-driven disaster reports while successfully mitigating common issues such as disaster bias and hallucinations. Ultimately, this work narrows the semantic gap between raw remote sensing data and actionable decision support, providing a robust foundation for the next generation of all weather, interactive humanitarian assistance systems.
\bibliographystyle{IEEEtran}
\bibliography{main}
\end{document}